\documentclass{article}

\usepackage[nonatbib,final]{neurips_2024}

\usepackage[utf8]{inputenc} 
\usepackage[T1]{fontenc}    
\usepackage{hyperref}       
\usepackage{url}            
\usepackage{booktabs}       
\usepackage{amsfonts}       
\usepackage{nicefrac}       
\usepackage{microtype}      
\usepackage[dvipsnames]{xcolor}         
\usepackage{bm}
\usepackage{mathtools}
\usepackage{cleveref}
\usepackage{tabularx}
\usepackage{algorithm}
\usepackage{algpseudocode}
\usepackage{marginnote}
\usepackage[numbers]{natbib}

\usepackage{tikz}
\usetikzlibrary{positioning}
\usetikzlibrary{shapes.geometric}
\usetikzlibrary{arrows.meta}
\usetikzlibrary{shapes.multipart}

\title{An Introduction to Discrete Variational Autoencoders}

\newcommand*{\ov}[1]{%
  \overline{\mbox{#1}\raisebox{2.5mm}{}}%
}
\newcommand\margincol{Gray}

\author{%
  Alan Jeffares \\
  University of Cambridge \\
  \texttt{aj659@cam.ac.uk} \\
  \And
  Liyuan Liu \\
  Microsoft Research \\
  \texttt{lucliu@microsoft.com}
}

\DeclareMathOperator*{\argmax}{arg\,max}

\begin{document}

\maketitle
\setlength\marginparwidth{3.2cm}

\begin{abstract}
  Variational Autoencoders (VAEs) are well-established as a principled approach to probabilistic unsupervised learning with neural networks. Typically, an encoder network defines the parameters of a Gaussian distributed latent space from which we can sample and pass realizations to a decoder network. This model is trained to reconstruct its inputs and is optimized through the \textit{evidence lower bound}. In recent years, \textit{discrete} latent spaces have grown in popularity, suggesting that they may be a natural choice for many data modalities (e.g. text). In this tutorial, we provide a rigorous, yet practical, introduction to discrete variational autoencoders -- specifically, VAEs in which the latent space is made up of latent variables that follow a \textit{categorical distribution}. We assume only a basic mathematical background with which we carefully derive each step from first principles. From there, we develop a concrete training recipe and provide an example implementation, hosted at \url{www.github.com/alanjeffares/discreteVAE}.
\end{abstract}

\section{Introduction}
The Variational Autoencoder (VAE) was formally introduced in the concurrent works of \cite{kingma2013auto} \& \cite{rezende2014stochastic} building upon earlier ideas such as the autoencoder architecture \cite{kramer1991nonlinear} and a stochastic \textit{recognition model} for modeling an encoder \cite{dayan1995helmholtz}. VAEs can be viewed as an unsupervised technique for learning a probabilistic, latent representation of the input distribution -- often in a much lower dimensional space. This is jointly optimized by an encoder and decoder network that together maximize a lower bound of the likelihood of the observed data (see \Cref{fig:vae}). The latent probability distribution is generally selected to be a parametric distribution from which we can easily sample and, thereby, generate new, unseen data. A popular choice is the Gaussian distribution which has the additional benefit of being amenable to the \textit{reparameterization trick} which replaces sampling $y \sim \mathcal{N}(\mu, \sigma)$ with the equivalent calculation $y = \mu + \sigma \cdot \epsilon$ where $\epsilon \sim \mathcal{N}(0, 1)$. This allows us to apply standard backpropogation while sidestepping the need to calculate gradients through the sampling process. 

Despite the qualities of the original Gaussian formulation of the VAE, in recent years there has been significant interest in discrete VAEs -- that is, variational encoders with a non-continuous latent space. It has been argued that a discrete representation may be a more appropriate inductive bias for many practical modalities such as text or images with several approaches introduced to develop discrete VAEs that are competitive with their continuous counterparts \citep[e.g.][]{rolfe2017discrete, vahdat2018dvae++}. A particularly successful extension has been \textit{vector quantized} VAEs \citep{van2017neural, razavi2019generating} that map inputs to discrete vector embeddings helping to avoid the problem of posterior collapse. Elsewhere, learning sparsely activated autoencoder representations has emerged as a popular approach to interpreting the activations of large language models requiring a partial discretization during learning \citep{gao2024scaling, lieberum2024gemma}. More broadly, discrete VAEs have acted as a testbed for developing better methods for approximating gradients when sampling from discrete latent variables.\footnote{This means solving $\nabla_{\bm{\psi}} \mathbb{E}_{x \sim \text{categorical}(\bm{\psi})}[f(x)]$, where the reparameterization trick cannot be used as we are sampling from the discrete categorical distribution rather than a Gaussian.} Here, significant progress has been made \citep[e.g.][]{jang2017categorical, liu2023bridging} resulting in advancements in a wide range of applications that require a similar sampling (e.g. mixture-of-expert layers in large language models \cite{liu2024grin}).

In this tutorial, we describe a canonical discrete VAE where the latent space consists of several independent categorical variables -- this is distinct from the many excellent existing tutorials and textbooks focused on the classic Gaussian VAE \citep[e.g.][]{kingma2019introduction, odaibo2019tutorial, doersch2016tutorial, ghojogh2021factor, prince2023understanding, murphy2023probabilistic, bishop2023deep}. While the broad idea remains unchanged, the ultimate implementation of the discrete VAE looks quite different from that of a Gaussian VAE meaning that some care must be taken in its derivation. We have endeavored to make this tutorial \textit{clear} (in terms of notation, illustration, etc.), \textit{complete} (each step of the derivation is detailed), and \textit{self-contained} (no prior knowledge of the VAE is assumed). In \Cref{sec:preliminaries}, we revise the necessary background including some basic probability identities and measures, a refresher on discrete probability distributions, maximum likelihood estimation, and Monte Carlo sampling. In \Cref{sec:introVAE,sec:vaeprob}, we describe an arbitrary VAE mechanistically and probabilistically. In \Cref{sec:vaegrads}, we derive the general form of a VAEs gradients. In \Cref{sec:thediscretevae}, we introduce the special case of a discrete VAE and in \Cref{sec:discretevaegrads} we derive its gradients concretely. Finally, in \Cref{sec:summary} we consolidate everything into a straightforward training recipe and provide a minimalist implementation using the \texttt{PyTorch} library.

\section{Preliminaries} \label{sec:preliminaries}
In this section, we provide a brief refresher on many of the technical details used throughout this tutorial. We assume only an introductory knowledge of deep learning, calculus, and probability. 

\subsection{Notation}
Scalar-valued variables will be denoted in regular font (i.e. $x$) while vectors, matrices, and concatenations of vectors are denoted in bold-face (i.e. $\mathbf{x}$). These objects are then indexed by subscripts alone in the case of vectors (i.e. $\mathbf{x}_i$) and superscripts too in the other cases (i.e. $\mathbf{x}_i^{(j)}$). For probability distributions we use a calligraphic font, and for an arbitrary distribution we will use $\mathcal{D}$. We will use the shorthand $x \sim \mathcal{D}$ to say that $x$ is a random variable or a realization of a random variable whose distribution follows $\mathcal{D}$. Often in machine learning, we parameterize a \textit{family} of distributions with a vector of learnable parameters $\bm{\psi}$ (e.g. the mean and covariance of a Gaussian). In this case, it is convenient to refer to the probability density/mass function of the distribution with the letter $p$ or $q$. Then we can refer to the evaluation of $x$ for a set of parameters $\bm{\psi}$ as $p_{\bm{\psi}}(x)$. When we place this expression in the subscript of an expectation (i.e., $\mathbb{E}_{p_{\bm{\psi}}(x)}[f(x)]$), this is an informal shorthand notation meaning that the expectation is taken with respect to a random variable whose distribution follows this family with parameters $\bm{\psi}$. Thus, $\mathbb{E}_{p_{\bm{\psi}}(x)}[f(x)] = \int_x p_{\bm{\psi}}(x)f(x) dx$.

\subsection{Probabilities and Information Measures}\label{sec:probabilities}

We will be using probabilistic terminology throughout and will therefore refer to the \textit{probability of some event} $A$ as $p(A)$. We will also denote the \textit{joint probability} of two events $A$ and $B$ as $p(A,B)$. If these two events are not independent (i.e. $p(A,B) \neq p(A)p(B)$) then we might be interested in the probability of event $A$ conditioned on event $B$ which is the \textit{conditional probability} and is denoted $p(A|B)$. Then the relationship between the joint and conditional probabilities is given by
\begin{equation}
    p(A,B) = p(A|B)p(B) = p(B|A)p(A). \label{eqn:ruleofconditionalprob}
\end{equation}
Sometimes we will wish to enumerate over all $k$ potential events. For a finite sample space, this amounts to taking the sum where the axioms of probability state that $\sum_{i=1}^kp(A_i) = 1$ and $p(A_i) \geq 0\; \forall\; i$. This leads us to the \textit{law of total probability} which states that
\begin{equation}
    \sum_{i=1}^kp(A,B_i) = p(A|B_i)p(B_i) = p(A).
\end{equation}
This is often referred to as \textit{marginalizing} over $B$. In the case of an arbitrary, potentially infinite sample space we will generalize the sum to an integral, but the same rules apply. Finally, we generally don't know the true probability of any given event and will attempt to model it somehow. We will often do so with a parameterized model such as a neural network. We will generally make this explicit by including the parameter vector of the model, say $\bm{\psi}$, in the subscript of the probability. Therefore, the probability of event $A$ under a given model parameterized by $\bm{\psi}$ is given by $p_{\bm{\psi}}(A)$.

Sometimes we will wish to measure the difference\footnote{Strictly speaking $D_{\text{KL}}$ measures \textit{difference} not \textit{distance} as it is not symmetric (i.e. $D_{\text{KL}}(q||p) \neq D_{\text{KL}}(p||q)$).} between two probability distributions -- i.e. how much they \textit{diverge}. A common measure is given by the \textit{Kullback–Leibler} or \textit{KL divergence}. For two probability distributions $p$ and $q$, this is given by
\begin{equation}
    D_{\text{KL}}\left(q(A)||p(A)\right) \coloneq \int q(A)\log \left[\frac{q(A)}{p(A)}\right] dA = -\int q(A)\log \left[\frac{p(A)}{q(A)}\right] dA. \label{eqn:KLdivergence}
\end{equation}
Another important idea from information theory is that of \textit{entropy} which quantifies the uncertainty associated with a probability distribution's potential states. This is given by
\begin{equation}
    \text{H}(p(A)) \coloneq - \int p(A) \log p(A) dA. 
\end{equation}
Finally, we will also need the concept of \textit{cross-entropy} which is another popular measure of the difference between two probability distributions given by 
\begin{equation}
    \text{H}(p(A), q(A)) \coloneq - \int p(A) \log q(A) dA.
\end{equation}

It is important to reiterate that the integrals in KL-divergence, entropy, and binary cross entropy all become sums in the case of a discrete probability distribution -- as we will eventually use in this tutorial. Notation can often get a little muddied when comparing these expressions to what is typically described in library implementations. This is because these implementations are typically expecting vector inputs. Therefore, we will also introduce more applied definitions of entropy and cross-entropy that agree with popular library implementations and expect discrete vector inputs $\mathbf{a} = [a_1, \ldots, a_k]$ and $\mathbf{b} = [b_1, \ldots, b_k]$\footnote{These vectors represent probabilities such that $\sum_{i=1}^ka_i = 1$ and $a_i \geq 0 \; \forall \; i$.}. Then \textit{entropy} can be expressed as
\begin{equation}
    \text{Entropy}(\mathbf{a}) \coloneq - \sum_{i=1}^k a_i \log a_i.\label{eqn:Entropy}
\end{equation}
Similarly, \textit{cross-entropy} is written as
\begin{equation}
    \text{CE}(\mathbf{a}, \mathbf{b}) \coloneq - \sum_{i=1}^k a_i \log b_i. 
\end{equation}
In the case of only two possible levels of the distribution (i.e. $k = 2$) then this reduces to \textit{binary cross-entropy} which we might wish to write explicitly as
\begin{equation}
\begin{split}
    \text{BCE}(\mathbf{a}, \mathbf{b}) \coloneq& - a_1 \log b_1 - a_2 \log b_2 \\
    =& - a_1 \log b_1 - (1 - a_1) \log (1 - b_1).
\end{split} \label{eqn:BCE}
\end{equation}
Often we will want to apply these functions over a batch of vectors in parallel (e.g. multiple latent variables). To make this neater, we will introduce one further shorthand notation. That is, if denote the concatenation of $m$ vectors as $\bm{\alpha} = [\mathbf{a}^{(1)}, \ldots, \mathbf{a}^{(m)}]$ and $\bm{\beta} = [\mathbf{b}^{(1)}, \ldots, \mathbf{b}^{(m)}]$, then we define \textit{aggregate entropy} as 
\begin{equation}
\ov{\text{Entropy}}(\bm{\alpha}) \coloneq \sum_{j=1}^m \text{Entropy}(\textbf{a}^{(j)}).\label{eqn:aggEntropy}    
\end{equation}
Similarly, we write \textit{aggregate binary cross-entropy} as
\begin{equation}
\ov{\text{BCE}}(\bm{\alpha}, \bm{\beta}) \coloneq \sum_{j=1}^m \text{BCE}(\textbf{a}^{(j)}, \textbf{b}^{(j)}).\label{eqn:aggBCE}    
\end{equation}

\subsection{Discrete Probability Distributions}

We will want to be able to describe the \textit{distribution} of potential outcomes for a discrete random variable $X$. Therefore, two particular discrete probability distributions will be useful to us later.

Firstly, the \textit{Bernoulli distribution} which describes an event with two possible outcomes (e.g. a coin flip). In this case, we say
\begin{equation}
    X \sim \text{Bernoulli}(p),
\end{equation}
where $p$ is a (typically learned) parameter that determines the probability of one outcome (thus $1 - p$ is the probability of the other). Then, the \textit{probability mass function} informs us of the probability that $X$ will take a given value $x$ -- i.e. $f_X(x) = p(X=x)$.

The \textit{probability mass function} of the Bernoulli distribution is given by
\begin{equation} \label{eqn:bernoullidistribution}
f_X(x) = \begin{cases}
    p &\;\;\text{if}\;\; x = 1 \\
    1 - p &\;\;\text{if}\;\; x = 0.
\end{cases}
\end{equation}
Note that this can be equivalently written as $f_X(x) = p^x (1-p)^{(1-x)}$.

We will also be interested in the \textit{categorical distribution} which simply extends the the Bernoulli distribution for $>2$ outcomes. Now, when a random variable $X$ follows the categorical distribution with $k$ possible outcomes, we say
\begin{equation}
    X \sim \text{Cat}(\textbf{p}),
\end{equation}
where $\mathbf{p} = [p_1, \ldots, p_k]$ is a parameter vector denoting the probability of each of the $k$ potential outcomes. In this more general case the probability mass function is given by
\begin{equation}
    f_X(x) = \begin{cases}
    p_1 &\;\;\text{if}\;\; x = 1 \\
    \;\vdots &\;\;\;\;\;\;\;\; \vdots \\
    p_k &\;\;\text{if}\;\; x = k. 
\end{cases}\label{eqn:categoricaldist}
\end{equation}
This probability mass function can also be written more conveniently but requires us to introduce the notation of the \textit{Iverson bracket},
\begin{equation} \label{eqn:iverson}
[\text{Statement}] = \begin{cases}
    1 &\;\;\text{if}\; \text{Statement is true}  \\
    0 &\;\;\text{Otherwise} .
\end{cases}
\end{equation}
Now we can write $f_X(x) = \sum_{i=1}^k[x=i] \cdot p_i$ for a neater version of the categorical distributions probability mass function. 

\subsection{Maximum Likelihood Estimation}\label{sec:maximumlikelihood}

Whenever we specify a parametric model that can estimate $p_{\bm{\psi}}(X)$, a common way of selecting the ``best'' parameters of that model $\bm{\hat{\psi}}$ is by finding the parameters that maximize the joint probability of the observed data under that model. That is, for some observed data $\{\mathbf{x}_i \}_{i=1}^n$, we seek the parameters that maximize the likelihood
\begin{equation}
    \bm{\hat{\psi}} = \argmax_{\bm{\psi}}\mathcal{L}(\bm{\psi}) = \argmax_{\bm{\psi}} \prod_{i=1}^n p_{\bm{\psi}}(x_i).
\end{equation}
This solution is known as the \textit{maximum likelihood estimator}. Often the product in this expression can be problematic so we instead consider the \textit{log-likelihood}, which has an identical solution $\bm{\hat{\psi}}$ and is generally easier to work with,
\begin{equation}
    \bm{\hat{\psi}} = \argmax_{\bm{\psi}}\ell(\bm{\psi}) = \argmax_{\bm{\psi}} \sum_{i=1}^n \log p_{\bm{\psi}}(x_i).
\end{equation}
Therefore, when designing our statistical models, we will often spend considerable time thinking about how, for some arbitrary data point $x$, we can maximize $\log p_{\bm{\psi}}(x)$. 

\subsection{Monte Carlo Sampling}\label{sec:montecarlo}
To optimize our models, we will generally require gradients of an expectation of the form
\begin{equation}\label{eqn:montecarlo}
    \nabla_{\bm{\phi}}\mathbb{E}_{\mathbf{x} \sim \mathcal{D}}[f_{\bm{\phi}}(\mathbf{x})].
\end{equation}
Where $\mathcal{D}$ is some distribution that doesn't depend on $\bm{\phi}$. Typically we will not be able to calculate the gradients for every element contributing to the expectation (i.e. for every element in the integral or sum). However, it may be practical to calculate the gradients for a single sample
\begin{equation}
    \nabla_{\bm{\phi}}f_{\bm{\phi}}(\mathbf{x}).
\end{equation}
This is known as a \textit{Monte Carlo estimate} and provides us with an unbiased estimate of the true gradient. By this, we (informally) mean that, as our number of samples approaches infinity, the average of the gradient samples is equal to the gradient of the expectation,
\begin{equation}
    \sum_{i = 1}^n\nabla_{\bm{\phi}}f_{\bm{\phi}}(\mathbf{x}_i) \to \nabla_{\bm{\phi}}\mathbb{E}_{\mathbf{x} \sim \mathcal{D}}[f_{\bm{\phi}}(\mathbf{x})] \:\: \text{as } n \to \infty.
\end{equation}
The details of this statement rely on the \textit{law of large numbers} and throughout this tutorial we will use the notation ``$\overset{\mathrm{mc}}{\approx}$'' to refer to cases of gradients being obtained as a Monte Carlo approximation. Specifically, we will say
\begin{equation}
    \nabla_{\bm{\phi}}\mathbb{E}_{\mathbf{x} \sim \mathcal{D}}[f_{\bm{\phi}}(\mathbf{x})] \overset{\mathrm{mc}}{\approx} \nabla_{\bm{\phi}}f_{\bm{\phi}}(\mathbf{x}).
\end{equation}
This is exactly the approximation used in \textit{stochastic gradient descent} where we are unable to calculate the exact gradients with respect to the entire population $\mathcal{D}$, and we can instead use a sample batch from that distribution to approximate it.

\subsection{The Log-Derivative Trick}\label{sec:logderivative}

Not all expectations are immediately amenable to Monte Carlo sampling like \Cref{eqn:montecarlo}. Consider the case where the function itself is independent of our parameters of interest but the probability distribution over which we are taking the expectation is dependent on those parameters i.e.
\begin{equation}
    {\nabla}_{\bm{\psi}} \mathbb{E}_{p_{\bm{\psi}}(\mathbf{x})}[f(\mathbf{x})]. 
\end{equation}
Now, we cannot simply move the gradient inside the expectation and apply Monte Carlo sampling, as the expectation itself is a function of $\bm{\psi}$. To make progress we can apply what is known as the \textit{log-derivative trick}.\footnote{Also known as the \textit{score function estimator} or the \textit{REINFORCE trick}.} This uses the fact that $p_{\bm{\psi}}(\mathbf{x}){\nabla}_{\bm{\psi}} \log p_{\bm{\psi}}(\mathbf{x}) = {\nabla}_{\bm{\psi}} \, p_{\bm{\psi}}(\mathbf{x})$ to show
\begin{align}
    {\nabla}_{\bm{\psi}} \mathbb{E}_{p_{\bm{\psi}}(\mathbf{x})}[f(\mathbf{x})] &= {\nabla}_{\bm{\psi}} \int_{\mathbf{x}} p_{\bm{\psi}}(\mathbf{x})f(\mathbf{x}) d\mathbf{x} \\
    &= \int_{\mathbf{x}} f(\mathbf{x}) {\nabla}_{\bm{\psi}} \, p_{\bm{\psi}}(\mathbf{x}) d\mathbf{x} \\
    &= \int_{\mathbf{x}} f(\mathbf{x}) \, p_{\bm{\psi}}(\mathbf{x}) \, {\nabla}_{\bm{\psi}} \log p_{\bm{\psi}}(\mathbf{x}) d\mathbf{x} \\
    &= \mathbb{E}_{p_{\bm{\psi}}(\mathbf{x})}[f(\mathbf{x}) \, {\nabla}_{\bm{\psi}} \log p_{\bm{\psi}}(\mathbf{x})]. \label{eqn:reinforce}
\end{align}
Now, in \Cref{eqn:reinforce}, the gradient only appears inside the expectation, and we can obtain a Monte Carlo estimate as before where
\begin{equation}
    {\nabla}_{\bm{\psi}} \mathbb{E}_{p_{\bm{\psi}}(\mathbf{x})}[f(\mathbf{x})] \overset{\mathrm{mc}}{\approx} f(\mathbf{x}) \, {\nabla}_{\bm{\psi}} \log p_{\bm{\psi}}(\mathbf{x}).
\end{equation}

\section{What is a Variational Autoencoder Mechanistically?} \label{sec:introVAE}
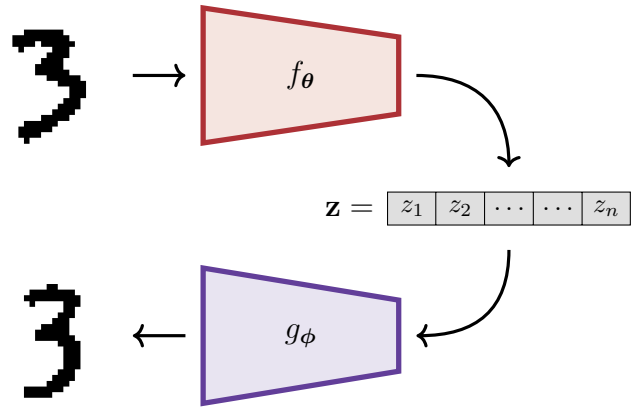
\begin{figure}[h]
\centering
\begin{tikzpicture}[align=left]
\tikzset{trapezium stretches=true}
\node (inputAE) {\includegraphics[width=.15\textwidth]{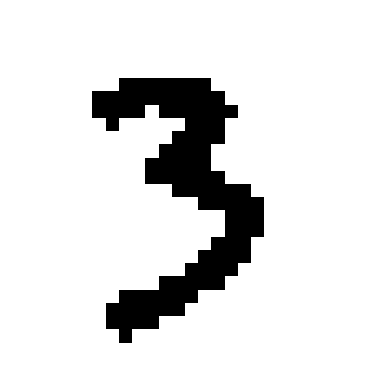}};
\node[trapezium, fill=Maroon!10, minimum width=1.8cm, minimum height=2.6cm, trapezium angle = 70, shape border rotate = 270, draw=Maroon, line width=0.7mm] (encoderAE) [right=0.8cm of inputAE] {\large $f_{\bm{\theta}}$};
\draw[->, very thick, shorten >= 0.2cm, shorten <= -0.1cm] (inputAE.east) to (encoderAE.west);
\node[rectangle split, rectangle split parts=5,  rectangle split horizontal, draw, text width=0.4cm, align=center, fill=gray!25,] (zAE) [below right= 0.9cm and 0.5cm of encoderAE] {$z_1$\nodepart{two}$z_2$\nodepart{three}$\ldots$\nodepart{four}$\ldots$\nodepart{five}$z_n$};
\node (znorthAE) at (zAE.north) {};
\draw [->, very thick, shorten >= 0.2cm, shorten <= 0.2cm, distance=1cm] (encoderAE.east) to[out=0,in=90] (znorthAE);
\node[trapezium, fill=RoyalPurple!10, minimum width=1.8cm, minimum height=2.6cm, trapezium angle = 70, shape border rotate = 270, draw=RoyalPurple, line width=0.7mm] (decoderAE) [below=2cm of encoderAE] {\large$g_{\bm{\phi}}$};
\node (zsouthAE) at (zAE.south) {};
\node [left = 0.05cm of zAE.west] {\large$\mathbf{z} = $};
\draw [->, very thick, shorten >= 0.2cm, shorten <= 0.2cm, distance=1cm] (zsouthAE) to[out=270,in=0] (decoderAE.east);
\node (outputAE) [left=0.8cm of decoderAE] 
{\includegraphics[width=.15\textwidth]{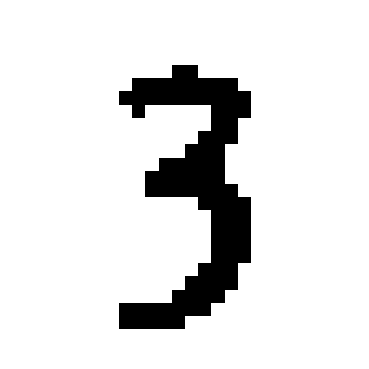}};
\draw[->, very thick, shorten <= 0.2cm, shorten >= -0.1cm] (decoderAE.west) to (outputAE.east);
\end{tikzpicture}
\caption{\textbf{An Autoencoder.} The encoder network $f_{\bm{\theta}}$ maps to a latent representation $\mathbf{z}$ which is decoded by a second network, $g_{\bm{\phi}}$, attempting to reproduce the original input.}
\label{fig:autoencoder}
\end{figure}
An autoencoder is a neural network that projects an input to a (typically) lower-dimensional latent representation and then attempts to reconstruct that same input from its latent representation -- effectively attempting to act as the identity function despite squeezing each input through an information bottleneck.
This can be seen as a direct form of \textit{lossy compression} where we attempt to find a more compact representation of the data that still encodes its most essential information. This concept has broad practical applications in a deep learning context (e.g. data generation and anomaly detection) where it has been highly influential. The network is conceptually subdivided into an \textit{encoder} and \textit{decoder} around the latent representation, which act as the compressor and reconstructor, respectively. This architecture is illustrated in \Cref{fig:autoencoder}. 

A variational autoencoder (VAE) is a probabilistic instantiation of the general autoencoder framework that learns to produce a \textit{probability distribution} in the latent state rather than just a single deterministic vector. This can be implemented by specifying a particular probability distribution over the latent variables (e.g. a Gaussian) and asking the model to learn the parameters of that distribution instead (e.g. mean and covariance).

\begin{figure}
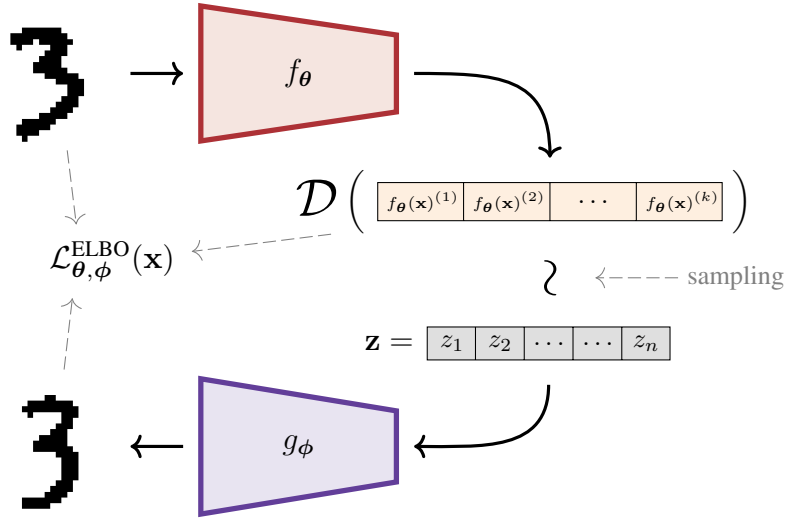

\centering
\begin{tikzpicture}[align=left]
\tikzset{trapezium stretches=true}
\node (inputAE) {\includegraphics[width=.15\textwidth]{figures/input_0.png}};
\node[trapezium, fill=Maroon!10, minimum width=1.8cm, minimum height=2.6cm, trapezium angle = 70, shape border rotate = 270, draw=Maroon, line width=0.7mm] (encoderAE) [right=0.8cm of inputAE] {\large$f_{\bm{\theta}}$};
\draw[->, very thick, shorten >= 0.2cm, shorten <= -0.1cm] (inputAE.east) to (encoderAE.west);
\node[rectangle split, rectangle split parts=4,  rectangle split horizontal, draw, text width=0.9cm, align=center, fill=Apricot!20,] (paramsAE) [below right= 0.8cm and 0.4cm of encoderAE] {\tiny$f_{\bm{\theta}}(\mathbf{x})^{(1)}$\nodepart{two}\tiny$f_{\bm{\theta}}(\mathbf{x})^{(2)}$\nodepart{three}$\ldots$\nodepart{four}\tiny$f_{\bm{\theta}}(\mathbf{x})^{(k)}$};
\node (left-paren) [left = -0cm of paramsAE] {\raisebox{-1mm}{\scalebox{2}{$\mathcal{D}$}}\;$\left(\vphantom{\text{\huge X}}\right.$};
\node (right-paren) [right = -0cm of paramsAE] {$\left)\vphantom{\text{\huge X}}\right.$};
\node[rectangle split, rectangle split parts=5,  rectangle split horizontal, draw, text width=0.4cm, align=center, fill=gray!25,] (zAE) [below left= 1.4cm and -3.9cm of paramsAE] {$z_1$\nodepart{two}$z_2$\nodepart{three}$\ldots$\nodepart{four}$\ldots$\nodepart{five}$z_n$};
\node (znorthAE) at (paramsAE.north) {};
\draw [->, very thick, shorten >= 0.2cm, shorten <= 0.2cm, distance=1.3cm] (encoderAE.east) to[out=0,in=90] (znorthAE);
\node[rotate=90] (sim) [above right = 0.25cm and -1.3cm of zAE] {\scalebox{2}{$\sim$}};
\node (simannotate) [right = 1.4cm of sim.south] {\textcolor{gray}{sampling}};
\draw[-{>[width=2.8mm, length=1.3mm]}, shorten <= 0cm, shorten >= 0.3cm, gray, dash pattern={on 5pt off 2pt}] (simannotate.west) to (sim.south);
\node[trapezium, fill=RoyalPurple!10, minimum width=1.8cm, minimum height=2.6cm, trapezium angle = 70, shape border rotate = 270, draw=RoyalPurple, line width=0.7mm] (decoderAE) [below=3.5cm of encoderAE] {\large$g_{\bm{\phi}}$};
\node (zsouthAE) at (zAE.south) {};
\node [left = 0.05cm of zAE.west] {\large$\mathbf{z} = $};
\draw [->, very thick, shorten >= 0.2cm, shorten <= 0.2cm, distance=1.1cm] (zsouthAE) to[out=270,in=0] (decoderAE.east);
\node (outputAE) [left=0.8cm of decoderAE] 
{\includegraphics[width=.15\textwidth]{figures/recon_0.png}};
\draw[->, very thick, shorten <= 0.2cm, shorten >= -0.1cm] (decoderAE.west) to (outputAE.east);
\node (ELBO) [above right = 0.9cm and -1.3cm of outputAE] {\textcolor{black}{\scalebox{1.2}{$\mathcal{L}^{\text{ELBO}}_{\bm{\theta}, \bm{\phi}}(\mathbf{x})$}}};
\draw[-{>[width=2.8mm, length=1.3mm]}, shorten <= -0.15cm, shorten >= 0.1cm, gray, dash pattern={on 5pt off 2pt}] (inputAE.282) to (ELBO.140);
\draw[-{>[width=2.8mm, length=1.3mm]}, shorten <= -0.15cm, shorten >= 0.1cm, gray, dash pattern={on 5pt off 2pt}] (outputAE.80) to (ELBO.217);
\draw[-{>[width=2.8mm, length=1.3mm]}, shorten <= -0.6cm, shorten >= 0.1cm, gray, dash pattern={on 5pt off 2pt}] (left-paren.220) to (ELBO);
\end{tikzpicture}
\caption{\textbf{A Variational Autoencoder.} Unlike the vanilla autoencoder, the encoder of this variational version outputs the \textit{parameters} of a probability distribution from which we sample and decode. The parameters are learned by maximizing the ELBO which provides a lower bound on the log-likelihood.}\label{fig:vae}
\end{figure}

Formally, for a p-dimensional input $\textbf{x} \in \mathbb{R}^p$, the encoder network will project to the parameters of a probability distribution $\mathcal{D}$ over the $n$ latent variables. These latent variables don't \textit{need} to be independent -- we could, for example, have a Gaussian distribution with a mean vector and a non-identity covariance matrix across the latent variables. However, for simplicity, we will treat each of the latent variables as being independent from this point onward. Therefore, each latent variable will be associated with its own probability distribution $\mathcal{D}^{(i)}$ with $k$ parameters\footnote{In the Gaussian example these would be mean and variance resulting in $k = 2$.}. If we stack the encoder network parameters into a vector $\bm{\theta}$ and express it as a function $f_{\bm{\theta}}:\mathbb{R}^p \to \mathbb{R}^{k\times n}$, then we can use our prespecified latent probability distribution $\mathcal{D}$ to obtain samples from the latent distribution $\mathbf{z} \sim \mathcal{D}(f_{\bm{\theta}}(\mathbf{x}))$ which means sampling $z_i \sim \mathcal{D}^{(i)}(f_{\bm{\theta}}(\mathbf{x})^{(i)})$ for each $i \in 1, \dots, n$. Note that the vector of samples $\mathbf{z} \in \mathbb{R}^n$ will be equal in length to the number of latent variables $n$. Similarly to the encoder, we can introduce a decoder network $g_{\bm{\phi}}:\mathbb{R}^n \to \mathbb{R}^p$ (parameterized by $\bm{\phi}$) that attempts to reconstruct the input from a sample as $\mathbf{\hat{x}} = g_{\bm{\phi}}(\mathbf{z})$. If our VAE is working well we should find that $\mathbf{\hat{x}} \approx \mathbf{x}$. This model is illustrated in \Cref{fig:vae}.

\section{What is a Variational Autoencoder Probabilistically?} \label{sec:vaeprob}

The previous description of a VAE is entirely mechanistic and does not provide us with much guidance on why this architecture might work or how we should optimize it. Making progress on this front requires us to view things through a \textit{probabilistic} lens. Let us start by considering the (log-)probability of an observation $\log p_{\bm{\theta}}(\mathbf{x})$ under a given set of parameters $\bm{\theta}$ (recall from \Cref{sec:maximumlikelihood} that this is the quantity we maximize in \textit{maximum likelihood estimation} making it a natural starting point). This is a little abstract as we have not specified how this could be modeled concretely, but let us stick with it for now (hint: it is not a coincidence that we are also denoting these parameters with $\bm{\theta}$). Fundamentally, the VAE is a \textit{latent variable model} meaning that we assume there to be some additional unobserved latent variables $\textbf{z}$ contributing to the model that we don't get to directly observe in the dataset. At least in theory, we can marginalize over these latent variables to make their contribution explicit.
\begin{equation} \label{eqn:marginal}
    p_{\bm{\theta}}(\mathbf{x}) = \int p_{\bm{\theta}}(\mathbf{x}, \mathbf{z}) d\mathbf{z}.
\end{equation}
In practice, this integral is not something we can typically compute and it is worth taking a brief aside to consider why we cannot simply specify some neural network here and find the parameters that maximize this quantity over a dataset (i.e. apply standard \textit{maximum likelihood estimation}). Attempting to do just this we can rewrite \Cref{eqn:marginal} such that 
\begin{equation}
    p_{\bm{\theta}}(\mathbf{x}) = \int p_{\bm{\theta}}(\mathbf{x} |\mathbf{z}) p_{\bm{\theta}}(\mathbf{z}) d\mathbf{z}.
\end{equation}
Modeling $p_{\bm{\theta}}(\mathbf{x} |\mathbf{z})$ with a neural network is reasonable and we can easily specify some prior $p_{\bm{\theta}}(\mathbf{z})$ but actually computing this integral is going to be intractable for any practical latent space\footnote{even if $\mathbf{z}$ is discrete and has $n$ dimensions (i.e. $\mathbf{z} = [z_1, z_2, \ldots, z_n]$) we would still need to calculate $ \sum_{{z}_1} \sum_{{z}_2} \ldots \sum_{{z}_n} p_{\bm{\theta}}(\mathbf{x}|\mathbf{z})p_{\bm{\theta}}(\mathbf{z})$ -- if each latent has $m$ values, this requires $m^n$ evaluations per input!}. Therefore, instead we will need to find some clever alternative. 

As a first step, we can now begin to decompose the log-likelihood in order to find a route towards finding some tractable objective that we can optimize. Let us begin by considering the rule of conditional probabilities from \Cref{eqn:ruleofconditionalprob},
\begin{equation}
    \begin{split}
        \log p_{\bm{\theta}}(\mathbf{x}) &= \log \left[ \frac{p_{\bm{\theta}}(\mathbf{x}, \mathbf{z})}{p_{\bm{\theta}}(\mathbf{z} | \mathbf{x})} \right].
    \end{split}
\end{equation}
Now without the problematic integral, the joint probability on the numerator $p_{\bm{\theta}}(\mathbf{x}, \mathbf{z})$ seems promising as something we can work with by specifying $p_{\bm{\theta}}(\mathbf{x} |\mathbf{z})$ and $p_{\bm{\theta}}(\mathbf{z})$. However, we are not so fortunate with the denominator. This is a posterior distribution of the latent variables under Bayes rule and would require $p_{\bm{\theta}}(\mathbf{x})$ itself in order to evaluate. In order to proceed we are going to separate these two terms and apply what will seem to be a very opaque trick. We will (twice) introduce a new arbitrary probability distribution $q_{\bm{\phi}}(\mathbf{z})$ over the latents parameterized by $\bm{\phi}$.
    \begin{align}
        \log p_{\bm{\theta}}(\mathbf{x})
        &= \mathbb{E}_{q_{\bm{\phi}}(\mathbf{z})}\left[\log p_{\bm{\theta}}(\mathbf{x}) \right] \label{eqn:liklihoodtrick1}\\
        &= \mathbb{E}_{q_{\bm{\phi}}(\mathbf{z})}\left[\log \left[ \frac{p_{\bm{\theta}}(\mathbf{x}, \mathbf{z})}{p_{\bm{\theta}}(\mathbf{z} | \mathbf{x})} \right] \right] \\
        &=  \mathbb{E}_{q_{\bm{\phi}}(\mathbf{z})}\left[\log \left[ \frac{p_{\bm{\theta}}(\mathbf{x}, \mathbf{z})}{q_{\bm{\phi}}(\mathbf{z})}\frac{q_{\bm{\phi}}(\mathbf{z})}{p_{\bm{\theta}}(\mathbf{z} | \mathbf{x})} \right]\right] \label{eqn:liklihoodtrick3}\\
        &= \mathbb{E}_{q_{\bm{\phi}}(\mathbf{z})}\left[\log \left[ \frac{p_{\bm{\theta}}(\mathbf{x}, \mathbf{z})}{q_{\bm{\phi}}(\mathbf{z})}\right]\right] + \mathbb{E}_{q_{\bm{\phi}}(\mathbf{z})}\left[\log \left[\frac{q_{\bm{\phi}}(\mathbf{z})}{p_{\bm{\theta}}(\mathbf{z} | \mathbf{x})} \right] \right].\label{eqn:liklihoodtrick4}
    \end{align}
Here $q_{\bm{\phi}}(\mathbf{z})$ was used twice: firstly, on line (\ref{eqn:liklihoodtrick1}) we took the expectation with respect to this distribution; secondly, on line (\ref{eqn:liklihoodtrick3}) we multiplied it in both the numerator and the denominator. So how has this helped us? Well, notice that the right-hand term on line (\ref{eqn:liklihoodtrick4}) -- containing the problematic posterior probability -- is exactly the KL-divergence (see \Cref{eqn:KLdivergence}) between the $q_{\bm{\phi}}(\mathbf{z})$ and $p_{\bm{\theta}}(\mathbf{z} | \mathbf{x})$ which, very importantly, is non-negative:
\begin{equation}
    \mathbb{E}_{q_{\bm{\phi}}(\mathbf{z})}\left[\log \left[\frac{q_{\bm{\phi}}(\mathbf{z})}{p_{\bm{\theta}}(\mathbf{z} | \mathbf{x})} \right] \right] = D_{\text{KL}}(q_{\bm{\phi}}(\mathbf{z})||p_{\bm{\theta}}(\mathbf{z} | \mathbf{x})) \geq 0.
\end{equation}
This tells us that the left-hand term of line (\ref{eqn:liklihoodtrick4}) is a \textit{lower bound} on the log-likelihood. This suggests a powerful way of maximizing the log-likelihood indirectly: if we can find the parameters $\bm{\theta}$ \& $\bm{\phi}$ that maximize this lower bound (without causing divergence on the other term) then this will also be a solution that comes close to maximizing the log-likelihood itself. Thus, the earlier trick of introducing $q_{\bm{\phi}}(\mathbf{z})$ has allowed us to sidestep the $p_{\bm{\theta}}(\mathbf{z} | \mathbf{x})$ term by ensuring its contribution to the objective is \textit{indirect} and now we can instead maximize the other, more workable term which is a lower bound on our ultimate objective. 

Just how ``close'' to the log-likelihood we will get is determined by how tight this lower bound actually is, so it is worth thinking about when this bound will be tight. Returning to \Cref{eqn:liklihoodtrick4}, it is clear that the bound is tightest when the KL divergence term is zero, and this occurs if and only if the two probability distributions are identical. Formally,
\begin{equation}
    D_{\text{KL}}(q_{\bm{\phi}}(\mathbf{z})||p_{\bm{\theta}}(\mathbf{z} | \mathbf{x})) = 0 \Leftrightarrow q_{\bm{\phi}}(\mathbf{z}) = p_{\bm{\theta}}(\mathbf{z} | \mathbf{x}).
\end{equation}
This motivates us to specify the (so far arbitrary) distribution $q_{\bm{\phi}}(\mathbf{z})$ to approximate $p_{\bm{\theta}}(\mathbf{z} | \mathbf{x})$ as closely as possible. Therefore, we introduce a dependence on $\mathbf{x}$ and henceforth refer to it as $q_{\bm{\phi}}(\mathbf{z}| \mathbf{x})$ aiming to find some reasonable approximation such that $q_{\bm{\phi}}(\mathbf{z}| \mathbf{x}) \approx p_{\bm{\theta}}(\mathbf{z} | \mathbf{x})$. 

Replacing $q_{\bm{\phi}}(\mathbf{z})$ with $q_{\bm{\phi}}(\mathbf{z}| \mathbf{x})$ in the left-hand term of \Cref{eqn:liklihoodtrick4} results in this lower bound taking a form known as the \textit{evidence lower bound} or the ELBO: 
\begin{equation}
    \mathcal{L}^{\text{ELBO}}_{\bm{\theta}, \bm{\phi}}(\mathbf{x}) \coloneq \mathbb{E}_{q_{\bm{\phi}}(\mathbf{z}|\mathbf{x})}\left[\log \left[ \frac{p_{\bm{\theta}}(\mathbf{x}, \mathbf{z})}{q_{\bm{\phi}}(\mathbf{z}|\mathbf{x})}\right]\right]. \label{eqn:ELBO} \marginnote{\small \textcolor{\margincol}{The \textit{evidence lower bound} or ELBO.}}[-0.2cm]
\end{equation}
Putting this all together, we can again write our expression of the log-likelihood from \Cref{eqn:liklihoodtrick4} which becomes:
\begin{equation}\label{eqn:loglikihoodfull}
    \log p_{\bm{\theta}}(\mathbf{x}) = \mathcal{L}^{\text{ELBO}}_{\bm{\theta}, \bm{\phi}}(\mathbf{x}) + D_{\text{KL}}(q_{\bm{\phi}}(\mathbf{z}| \mathbf{x})||p_{\bm{\theta}}(\mathbf{z} | \mathbf{x})).
\end{equation}

Due to the non-negativity of the KL divergence, the ELBO now acts as a lower bound on the log-likelihood (i.e. $\log p_{\bm{\theta}}(\mathbf{x}) \geq \mathcal{L}^{\text{ELBO}}_{\bm{\theta}, \bm{\phi}}(\mathbf{x})$). Therefore, in order to \textit{implicitly} maximize the (intractable) log-likelihood, we can \textit{explicitly} maximize the (tractable) ELBO. This provides us with a concrete objective with which we can optimize the VAE. Before moving on, it is worth taking a moment to reflect on just how elegant this expression is. This is best illustrated by rearranging \Cref{eqn:loglikihoodfull} in terms of the ELBO.
\begin{equation}\label{eqn:loglikihoodrearrange}
    \mathcal{L}^{\text{ELBO}}_{\bm{\theta}, \bm{\phi}}(\mathbf{x}) = \log p_{\bm{\theta}}(\mathbf{x}) - D_{\text{KL}}(q_{\bm{\phi}}(\mathbf{z}| \mathbf{x})||p_{\bm{\theta}}(\mathbf{z} | \mathbf{x})).
\end{equation}
On the right-hand side, we now have the log-likelihood and the KL-divergence terms. When we maximize the ELBO, we are effectively doing two things: (1) We are \textit{minimizing} the KL divergence term which, as previously discussed, is achieved by making the approximate posterior, $q_{\bm{\phi}}(\mathbf{z}| \mathbf{x})$, a closer approximation of the true posterior, $p_{\bm{\theta}}(\mathbf{z} | \mathbf{x})$. Therefore, our model of the posterior distribution over the latent variables is improving; (2) We are \textit{maximizing} the log-likelihood of the data, $\log p_{\bm{\theta}}(\mathbf{x})$, ensuring that the generative model is better capturing the distribution of the observed data. In this way, the ELBO serves as a principled, interpretable, and computationally tractable objective for training any VAE. 


\section{The General Form of VAE gradients} \label{sec:vaegrads}
Given that we have derived a sensible optimization objective in the previous section, it is now time to start optimizing our parameters $\bm{\theta}$ and $\bm{\phi}$! Although these will be the parameters of a neural network, it is easier to keep these details abstract for now. The primary remaining obstacle is due to the fact that we will optimize these parameters using (stochastic) gradient descent and, therefore, need to calculate (or estimate) the gradients of the ELBO, $\mathcal{L}^{\text{ELBO}}_{\bm{\theta}, \bm{\phi}}(\mathbf{x})$. To do so, we will first further decompose this objective.
\begin{align}
    \mathcal{L}^{\text{ELBO}}_{\bm{\theta}, \bm{\phi}}(\mathbf{x}) &= \mathbb{E}_{q_{\bm{\phi}}(\mathbf{z}|\mathbf{x})}\left[\log \left[ \frac{p_{\bm{\theta}}(\mathbf{x}, \mathbf{z})}{q_{\bm{\phi}}(\mathbf{z}|\mathbf{x})}\right]\right] \label{eqn:elbodecomp1}\\
    &= \int q_{\bm{\phi}}(\mathbf{z}|\mathbf{x})\log \left[ \frac{p_{\bm{\theta}}(\mathbf{x}| \mathbf{z})p(\mathbf{z})}{q_{\bm{\phi}}(\mathbf{z}|\mathbf{x})}\right] d\mathbf{z} \label{eqn:elbodecomp2}\\
    &= \int q_{\bm{\phi}}(\mathbf{z}|\mathbf{x})\log \left[ \frac{p(\mathbf{z})}{q_{\bm{\phi}}(\mathbf{z}|\mathbf{x})}\right] d\mathbf{z}  + \int q_{\bm{\phi}}(\mathbf{z}|\mathbf{x})\log p_{\bm{\theta}}(\mathbf{x}| \mathbf{z})d\mathbf{z} \label{eqn:elbodecomp3}\\
    &= - D_{\text{KL}}(q_{\bm{\phi}}(\mathbf{z}|\mathbf{x})||p(\mathbf{z})) + \mathbb{E}_{q_{\bm{\phi}}(\mathbf{z}|\mathbf{x})}\left[ \log p_{\bm{\theta}}(\mathbf{x}| \mathbf{z}) \right]. \label{eqn:elbodecomp4}
\end{align}
On line (\ref{eqn:elbodecomp2}), we have expanded the expectation and applied \Cref{eqn:ruleofconditionalprob}. This requires us to introduce a prior $p(\mathbf{z})$ which we note \textit{does not} depend on parameters $\bm{\theta}$ as it is not something we will learn\footnote{In a standard, continuous VAE we would typically specify $\mathbf{z} \sim N(\mathbf{0}, \mathbf{I})$ for this prior which doesn't contain any learnable parameters.}. To make this explicit we do drop the subscript on this term. Then, on line (\ref{eqn:elbodecomp3}) we expand according to the laws of logs, resulting in two easily interpretable expressions. The expectation term asks the model to maximize the likelihood of the data given a latent representation, while the KL-divergence term acts as a regularization preventing too much divergence of our model of the latents from their prior. 

We are now ready to examine the general form of the gradients of both $\bm{\theta}$ and $\bm{\phi}$ which we denote as $\hat{\nabla}_{\bm{\theta}}$ and $\hat{\nabla}_{\bm{\phi}}$ respectively. Noting that KL-divergence term does not appear in $\hat{\nabla}_{\bm{\theta}}$ as it is constant with respect to $\bm{\theta}$, these gradients are given by: 
\begin{align}
    \hat{\nabla}_{\bm{\theta}} &\coloneq \nabla_{\bm{\theta}} \mathcal{L}^{\text{ELBO}}_{\bm{\theta}, \bm{\phi}}(\mathbf{x}) = {\nabla}_{\bm{\theta}} \mathbb{E}_{q_{\bm{\phi}}(\mathbf{z}|\mathbf{x})}\left[ \log p_{\bm{\theta}}(\mathbf{x}| \mathbf{z}) \right] \label{eqn:abstractthetagrads}\\
    \hat{\nabla}_{\bm{\phi}} &\coloneq \nabla_{\bm{\phi}} \mathcal{L}^{\text{ELBO}}_{\bm{\theta}, \bm{\phi}}(\mathbf{x}) = - {\nabla}_{\bm{\phi}} D_{\text{KL}}(q_{\bm{\phi}}(\mathbf{z}|\mathbf{x})||p(\mathbf{z})) + {\nabla}_{\bm{\phi}} \mathbb{E}_{q_{\bm{\phi}}(\mathbf{z}|\mathbf{x})}\left[ \log p_{\bm{\theta}}(\mathbf{x}| \mathbf{z}) \right]. \label{eqn:abstractphigrads} \marginnote{\small \textcolor{\margincol}{The general form of the gradients of any VAE.}}[-0.5cm]
\end{align}
Until now we have remained quite abstract on the specific form of the various $p$ and $q$ terms. Everything discussed thus far holds quite generally for \textit{any} VAE regardless of the form of the latent space $\mathbf{z}$ or inputs $\mathbf{x}$. Throughout the remainder of this tutorial, we will focus on the special case of a discrete latent space resulting in what is known as a \textit{discrete variational autoencoder}.

\section{The Discrete VAE} \label{sec:thediscretevae}
\begin{figure}
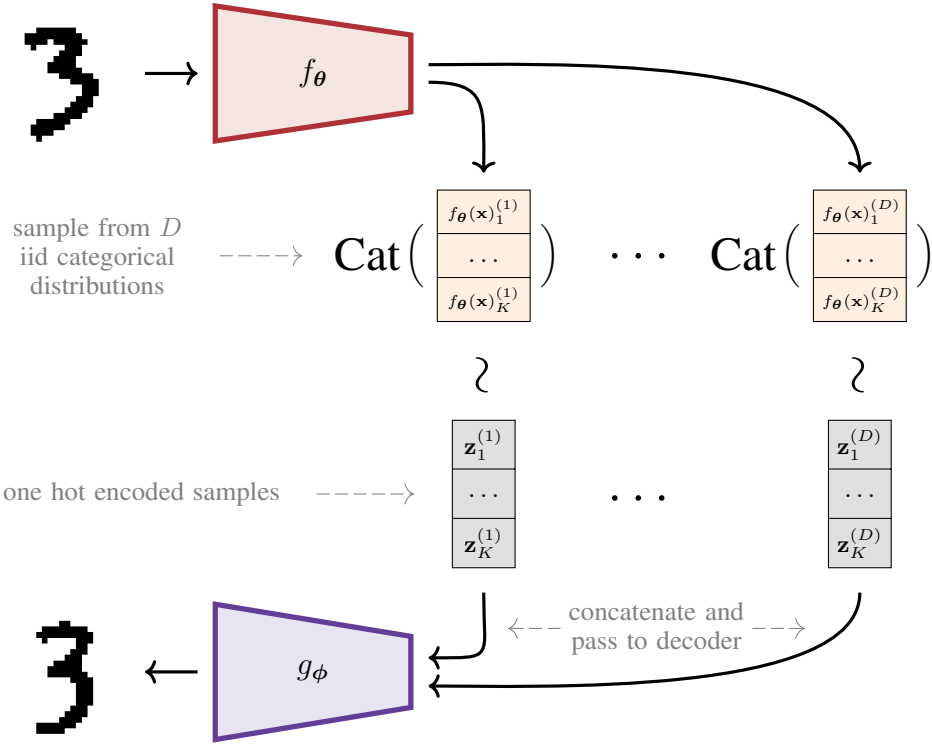

\centering
\begin{tikzpicture}[align=left]
\tikzset{trapezium stretches=true}
\node (inputAE) {\includegraphics[width=.15\textwidth]{figures/input_0.png}};
\node[trapezium, fill=Maroon!10, minimum width=1.8cm, minimum height=2.6cm, trapezium angle = 70, shape border rotate = 270, draw=Maroon, line width=0.7mm] (encoderAE) [right=0.8cm of inputAE] {\large$f_{\bm{\theta}}$};
\draw[->, very thick, shorten >= 0.2cm, shorten <= -0.1cm] (inputAE.east) to (encoderAE.west);
\node[rectangle split, rectangle split parts=3, draw, text width=1cm, align=center, fill=Apricot!20] (paramsAE) [below right= 0.9cm and 1cm of encoderAE] {\tiny$f_{\bm{\theta}}(\mathbf{x})^{(1)}_1$\nodepart{two}$\ldots${\tiny$\vphantom{)^{(1)}_K}$}\nodepart{three}\tiny$f_{\bm{\theta}}(\mathbf{x})^{(1)}_K$};
\node (left-paren) [left = -0.cm of paramsAE] {\raisebox{-1mm}{\scalebox{1.8}{$\text{Cat}$}}\,$\left(\vphantom{\text{\huge X}}\right.$};
\node (right-paren) [right = -0.cm of paramsAE] {$\left)\vphantom{\text{\huge X}}\right.$};
\node[rectangle split, rectangle split parts=3, draw, text width=0.6cm, align=center, fill=gray!25,] (zAE) [below = 1.3cm of paramsAE] {\small$\mathbf{z}^{(1)}_1$\nodepart{two}\raisebox{0.07cm}{$\ldots$}{\small$\vphantom{\mathbf{z}^{(1)}_K}$}\nodepart{three}{\small$\mathbf{z}^{(1)}_K$}};
\node[rotate=90] (sim) [above right = 0.2cm and -0.15cm of zAE] {\scalebox{2}{$\sim$}};

\node[rectangle split, rectangle split parts=3, draw, text width=1cm, align=center, fill=Apricot!20] (paramsAE2) [below right= 0.9cm and 6cm of encoderAE] {\tiny$f_{\bm{\theta}}(\mathbf{x})^{(D)}_1$\nodepart{two}$\ldots${\tiny$\vphantom{)^{(D)}_K}$}\nodepart{three}\tiny$f_{\bm{\theta}}(\mathbf{x})^{(D)}_K$};
\node (left-paren2) [left = -0.cm of paramsAE2] {\raisebox{-1mm}{\scalebox{1.8}{$\text{Cat}$}}\,$\left(\vphantom{\text{\huge X}}\right.$};
\node (right-paren2) [right = -0.cm of paramsAE2] {$\left)\vphantom{\text{\huge X}}\right.$};
\node[rectangle split, rectangle split parts=3, draw, text width=0.6cm, align=center, fill=gray!25,] (zAE2) [below = 1.3cm of paramsAE2] {\small$\mathbf{z}^{(D)}_1$\nodepart{two}\raisebox{0.07cm}{$\ldots$}{\small$\vphantom{\mathbf{z}^{(D)}_K}$}\nodepart{three}{\small$\mathbf{z}^{(D)}_K$}};
\node[rotate=90] (sim2) [above right = 0.2cm and -0.15cm of zAE2] {\scalebox{2}{$\sim$}};

\node (znorthAE) at (paramsAE.north) {};
\node (znorthAE2) at (paramsAE2.north) {};
\draw [->, very thick, shorten >= 0.1cm, shorten <= 0.2cm, distance=1cm] (encoderAE.355) to[out=0,in=90] (znorthAE);
\draw [->, very thick, shorten >= 0.1cm, shorten <= 0.2cm, distance=1.7cm] (encoderAE.5) to[out=0,in=90] (znorthAE2);

\node(upperdots) [left = 0.3cm of left-paren2] {\scalebox{2}{$\ldots$}};
\node(lowerdots) [below = 2.9cm of upperdots] {\scalebox{2}{$\ldots$}};

\node[trapezium, fill=RoyalPurple!10, minimum width=1.8cm, minimum height=2.6cm, trapezium angle = 70, shape border rotate = 270, draw=RoyalPurple, line width=0.7mm] (decoderAE) [below=6.5cm of encoderAE] {\large$g_{\bm{\phi}}$};
\node (zsouthAE) at (zAE.south) {};
\node (zsouthAE2) at (zAE2.south) {};
\draw [->, very thick, shorten >= 0.2cm, shorten <= 0.2cm, distance=1.1cm] (zsouthAE) to[out=270,in=0] (decoderAE.8);
\draw [->, very thick, shorten >= 0.2cm, shorten <= 0.2cm, distance=1.6cm] (zsouthAE2) to[out=270,in=0] (decoderAE.352);
\node (outputAE) [left=0.8cm of decoderAE] 
{\includegraphics[width=.15\textwidth]{figures/recon_0.png}};
\draw[->, very thick, shorten <= 0.2cm, shorten >= -0.1cm] (decoderAE.west) to (outputAE.east);

\node (catannotate) [left = 1.4cm of left-paren.west, text width = 3cm, align = center] {\textcolor{gray}{sample from $D$ iid categorical distributions}};
\draw[-{>[width=2.8mm, length=1.3mm]}, shorten <= 0cm, shorten >= 0.3cm, gray, dash pattern={on 5pt off 2pt}] (catannotate.east) to (left-paren.west);
\node (sampleannotate) [left = 2cm of zAE.west, text width = 4cm, align = center] {\textcolor{gray}{one hot encoded samples}};
\draw[-{>[width=2.8mm, length=1.3mm]}, shorten <= 0.2cm, shorten >= 0.5cm, gray, dash pattern={on 5pt off 2pt}] (sampleannotate.east) to (zAE.west);
\node (sampleannotate2) [below left = 1.1cm and -2.9cm of lowerdots, text width = 4cm, align = center] {\textcolor{gray}{concatenate and pass to decoder}};
\node [right = 0.85cm of sampleannotate2.west] (leftsource) {};
\node [left = 0.7cm of leftsource] (leftdestination) {};
\draw[-{>[width=2.8mm, length=1.3mm]}, shorten <= 0cm, shorten >= 0cm, gray, dash pattern={on 5pt off 2pt}] (leftsource) to (leftdestination);
\node [left = 0.85cm of sampleannotate2.east] (rightsource) {};
\node [right = 0.7cm of rightsource] (rightdestination) {};
\draw[-{>[width=2.8mm, length=1.3mm]}, shorten <= 0cm, shorten >= 0cm, gray, dash pattern={on 5pt off 2pt}] (rightsource) to (rightdestination);

\end{tikzpicture}
\caption{\textbf{The Discrete Variational Autoencoder.} The input is encoded into the parameters of $D$ categorical distributions; then, we sample one of the $K$ categories from each. These samples are passed back through the decoder that attempts to reproduce the original input.}\label{fig:discretevae}
\end{figure}

We are finally ready to apply the high-level theory established thus far towards a concrete instantiation of a \textit{discrete} VAE covering the necessary details required to train one of these models from scratch. Before proceeding, let us first formally describe what specific design choices of the general VAE framework are required to produce a discrete VAE. 

The key characteristic of a discrete VAE lies in the form of sample from its latent space $\mathbf{z}$. This space will consist of $D$ individual latent variables each of which has $K$ discrete categories. To practically implement this, we can represent each sample using a single one-hot-encoded vector per variable with a ``1'' at the index of the sampled category and ``0'' elsewhere. To be explicit we can denote $\mathbf{z} \in \{0, 1\}^{D\times K}$ such that $\sum_{k=1}^K\mathbf{z}^{(d)}_k = 1$ $\forall$ $d \in [1, D]$ with superscripts indexing latent variables and subscripts indexing their categories. A natural choice for the prior distribution $p(\mathbf{z})$ is the uniform categorical distribution over each latent dimension such that
\begin{equation}
    \mathbf{z}^{(d)} \overset{\mathrm{iid}}{\sim} \text{Cat}({\mathbf{K}^{-1}}). 
\end{equation}
Here, ${\mathbf{K}^{-1}} \coloneq [K^{-1}, \ldots, K^{-1}] \in \mathbb{R}^K$. A categorical distribution over the prior also implies a categorical distribution over the posterior $q_{\bm{\phi}}(\mathbf{z}|\mathbf{x})$, but with learned parameters $\bm{\phi}$ which we can model using a neural network $f_{\bm{\phi}}(\mathbf{x}): \mathcal{X}^P \to \Delta^{D\times K}$ where $\mathcal{X}$ denotes some arbitrary input space and $\Delta$ denotes a space of probabilities such that $\Delta \coloneq [0, 1]^{D\times K}$ where for any ${\mathbf{y}} \in \Delta$ it holds that $\sum_{k=1}^K{\mathbf{y}}^{(d)}_k = 1$ $\forall$ $d \in [1, D]$. Put more simply, $f_{\bm{\phi}}(\mathbf{x})$ denotes a neural network mapping from an input to the parameters of a categorical distribution which we will sample our discrete $\mathbf{z}$ from. Thus, during a forward pass we will obtain each dimension $d$ of our latent sample according to 
\begin{equation}
    \mathbf{z}^{(d)} \overset{\mathrm{iid}}{\sim} \text{Cat}(f_{\bm{\phi}}(\mathbf{x})^{(d)}). 
\end{equation}
This captures the \textit{encoder's} role in the discrete autoencoder, so all that remains is to specify the \textit{decoder}.  

Depending on what the inputs $\mathbf{x}$ are representing exactly, $p_{\bm{\theta}}(\mathbf{x}| \mathbf{z})$ will vary (i.e. what exactly is the space $\mathcal{X}$?). To keep things simple we will focus on the specific case of the MNIST dataset where each input represents a greyscale digit. We will further binarize each pixel to be black or white\footnote{This binarization is often not applied in practice at the expense of some theoretical rigor. However, for the Bernoulli distribution to be appropriate, we require $\mathbf{x} \in \{0,1\}^P$. See \cite{loaiza2019continuous} for an extended discussion on this point.} such that $\mathcal{X}^P \coloneq \{0,1\}^P$. Then, a natural prior over each pixel is the Bernoulli distribution which is a special case of the categorical distribution for just two categories. This also results in the posterior, $p_{\bm{\theta}}(\mathbf{x}| \mathbf{z})$, being Bernoulli which we can again parameterize using a neural network $g_{\bm{\theta}}(\mathbf{z}): \Delta^{D\times K} \to \mathcal{X}^P$. The decoder $g_{\bm{\theta}}(\mathbf{z})$ is therefore asked to reconstruct the original input probabilities $\mathbf{x}$ from the latent sample $\mathbf{z}$ such that 
\begin{equation}
    \mathbf{x}^{(p)} \overset{\mathrm{iid}}{\sim} \text{Bernoulli}(g_{\bm{\theta}}(\mathbf{z})^{(p)}). 
\end{equation}

\section{Deriving The Gradients and Loss of a Discrete VAE} \label{sec:discretevaegrads}
We can now derive the gradients $\hat{\nabla}_{\bm{\theta}}$ and $\hat{\nabla}_{\bm{\phi}}$ from \Cref{eqn:abstractthetagrads,eqn:abstractphigrads} for the discrete VAE described in the previous section. Let us begin with the decoder gradients $\hat{\nabla}_{\bm{\theta}}$ which are relatively straightforward. Since the expectation is over $q_{\bm{\phi}}(\mathbf{z}|\mathbf{x})$ which only depends on the $\bm{\phi}$ parameters we can bring the gradient \textit{inside} the expectation, giving
\begin{align}
    \hat{\nabla}_{\bm{\theta}} &= {\nabla}_{\bm{\theta}} \mathbb{E}_{q_{\bm{\phi}}(\mathbf{z}|\mathbf{x})}\left[ \log p_{\bm{\theta}}(\mathbf{x}| \mathbf{z}) \right] \\
    &= \mathbb{E}_{q_{\bm{\phi}}(\mathbf{z}|\mathbf{x})}\left[ {\nabla}_{\bm{\theta}} \log p_{\bm{\theta}}(\mathbf{x}| \mathbf{z}) \right].
\end{align}
This greatly simplifies our task as we can apply Monte Carlo sampling rather than having to calculate the expectation over the entire distribution $q_{\bm{\phi}}(\mathbf{z}|\mathbf{x})$ (see \Cref{sec:montecarlo} for details on Monte Carlo sampling). All that remains is to calculate ${\nabla}_{\bm{\theta}} \log p_{\bm{\theta}}(\mathbf{x}| \mathbf{z})$. We already know that each $\mathbf{x}^{(p)}$ is iid Bernoulli distributed over all the input features giving us
\begin{align}
    {\nabla}_{\bm{\theta}} \log p_{\bm{\theta}}(\mathbf{x}| \mathbf{z}) &= {\nabla}_{\bm{\theta}} \log \prod_{p=1}^P p_{\bm{\theta}}(\mathbf{x}^{(p)}| \mathbf{z}) \\
    &= \sum_{p=1}^P {\nabla}_{\bm{\theta}} \log p_{\bm{\theta}}(\mathbf{x}^{(p)}| \mathbf{z}).
\end{align}
Now, we recall from \Cref{eqn:bernoullidistribution} that the \textit{probability mass function} of the Bernoulli distribution is given by
\begin{equation}
    f_X(x) = p^x (1-p)^{(1-x)},
\end{equation}
Substituting this into the previous expression, we obtain
\begin{equation}
    \log p_{\bm{\theta}}(\mathbf{x}^{(p)}| \mathbf{z}) = \mathbf{x}^{(p)} \log p_{\bm{\theta}}(\mathbf{x}^{(p)} = 1| \mathbf{z}) + (1 - \mathbf{x}^{(d)})\log p_{\bm{\theta}}(\mathbf{x}^{(p)} = 0| \mathbf{z}).
\end{equation}
We can finally write our complete derivation of the gradients of $\bm{\theta}$ as 
\begin{align}
    \hat{\nabla}_{\bm{\theta}} &\overset{\mathrm{mc}}{\approx} \sum_{p=1}^P {\nabla}_{\bm{\theta}} \left( \mathbf{x}^{(p)} \log p_{\bm{\theta}}(\mathbf{x}^{(p)} = 1| \mathbf{z}) + (1 - \mathbf{x}^{(p)})\log p_{\bm{\theta}}(\mathbf{x}^{(p)} = 0| \mathbf{z}) \right) \label{eqn:discretethetagrads1}\\
    &= \sum_{p=1}^P {\nabla}_{\bm{\theta}} \left( \mathbf{x}^{(p)} \log g_{\bm{\theta}}(\mathbf{z})^{(p)} + (1 - \mathbf{x}^{(p)})\log (1 - g_{\bm{\theta}}(\mathbf{z})^{(p)}) \right) \label{eqn:discretethetagrads2} \\
    &= -{\nabla}_{\bm{\theta}} \; \sum_{p=1}^P\text{BCE}(g_{\bm{\theta}}(\mathbf{z})^{(p)}, \mathbf{x}^{(p)}) \label{eqn:discretethetagrads3} \\ 
    &= -{\nabla}_{\bm{\theta}} \; \ov{\text{BCE}}(g_{\bm{\theta}}(\mathbf{z}), \mathbf{x}) \label{eqn:discretethetagrads4} \marginnote{\small \textcolor{\margincol}{Derived gradients w.r.t. $\bm{\theta}$ from \cref{eqn:abstractthetagrads}.}}[-0.2cm]
\end{align}
On line (\ref{eqn:discretethetagrads1}) we denote the fact that we are now using an unbiased estimate of the true gradient through Monte Carlo sampling with the notation $\overset{\mathrm{mc}}{\approx}$. On line (\ref{eqn:discretethetagrads2}) we make the neural network's role explicit by including the decoder network $g_{\bm{\theta}}(\mathbf{z})$. Then on lines (\ref{eqn:discretethetagrads3}) \& (\ref{eqn:discretethetagrads4}) we simply notice that this expression takes the form of \textit{binary cross-entropy} as described in \Cref{sec:probabilities} and follow the notation of \Cref{eqn:aggBCE}. Calculating the gradients with respect to standard terms like binary cross-entropy is straightforward in any automatic differentiation library (e.g. \texttt{pytorch}), so this expression of the gradients is all we need to train these parameters of our discrete VAE.

We may now move on to the gradients with respect to $\bm{\phi}$ given by $\hat{\nabla}_{\bm{\phi}}$ from \Cref{eqn:abstractphigrads}. Let us start with the left-hand KL divergence term. In \Cref{sec:thediscretevae} we specified that each variable in $\mathbf{z}$ is iid according to the categorical distribution. Let us update \Cref{eqn:abstractphigrads} to reflect this as it was originally expressed for a single distribution over the entire latent space
\begin{align}
    -{\nabla}_{\bm{\phi}} D_{\text{KL}}(q_{\bm{\phi}}(\mathbf{z}|\mathbf{x})||p(\mathbf{z})) &= {\nabla}_{\bm{\phi}}\int q_{\bm{\phi}}(\mathbf{z}|\mathbf{x})\log \left[ \frac{p(\mathbf{z})}{q_{\bm{\phi}}(\mathbf{z}|\mathbf{x})}\right] d\mathbf{z} \\
    &= {\nabla}_{\bm{\phi}}\sum_{d=1}^D \sum_{k=1}^K q_{\bm{\phi}}(\mathbf{z}^{(d)}_k|\mathbf{x})\log \left[ \frac{p(\mathbf{z})_k}{q_{\bm{\phi}}(\mathbf{z}^{(d)}_k|\mathbf{x})}\right] \\
    &= -{\nabla}_{\bm{\phi}} \sum_{d=1}^D D_{\text{KL}}(q_{\bm{\phi}}(\mathbf{z}^{(d)}|\mathbf{x})||p(\mathbf{z})).\label{eqn:multikldiv1}
\end{align}
This leaves us with a sum of the KL divergences between the categorical posterior distribution of each latent variable and a uniform prior. Then, for each individual latent $\mathbf{z}^{(d)}$ we observe that
\begin{align}
    -D_{\text{KL}}(q_{\bm{\phi}}(\mathbf{z}^{(d)}|\mathbf{x})||p(\mathbf{z})) &= \sum_{k=1}^K q_{\bm{\phi}}(\mathbf{z}^{(d)}_k|\mathbf{x})\log \left[ \frac{p(\mathbf{z})_k}
    {q_{\bm{\phi}}(\mathbf{z}^{(d)}_k|\mathbf{x})}\right] \label{eqn:kldivderivation1}\\
    &= \sum_{k=1}^K q_{\bm{\phi}}(\mathbf{z}^{(d)}_k|\mathbf{x})\log {p(\mathbf{z})_k} - \sum_{k=1}^K q_{\bm{\phi}}(\mathbf{z}^{(d)}_k|\mathbf{x})\log {q_{\bm{\phi}}(\mathbf{z}^{(d)}_k|\mathbf{x})} \label{eqn:kldivderivation2}\\
    &= \log \frac{1}{K} \sum_{k=1}^K q_{\bm{\phi}}(\mathbf{z}^{(d)}_k|\mathbf{x}) - \sum_{k=1}^K q_{\bm{\phi}}(\mathbf{z}^{(d)}_k|\mathbf{x})\log {q_{\bm{\phi}}(\mathbf{z}^{(d)}_k|\mathbf{x})} \label{eqn:kldivderivation3}\\
    &= \log \frac{1}{K} - \sum_{k=1}^K f_{\bm{\phi}}(\mathbf{x})^{(d)}_k\log {f_{\bm{\phi}}(\mathbf{x})^{(d)}_k} \label{eqn:kldivderivation4}\\
    &= - \log K + \text{Entropy}(f_{\bm{\phi}}(\mathbf{x})^{(d)}).\label{eqn:kldivderivation5}
\end{align}
Lines (\ref{eqn:kldivderivation1}) \& (\ref{eqn:kldivderivation2}) simply expand the definition of KL-divergence and split into two terms. Line (\ref{eqn:kldivderivation3}) uses the fact that the parameters of the prior $p(\mathbf{z})$ are constant with each element given by $\frac{1}{K}$. Line (\ref{eqn:kldivderivation4}) utilizes the sum over all categories being equal to 1 and expresses all terms to make the role of the neural network explicit. Finally, on line (\ref{eqn:kldivderivation5}) we notice that the right-hand term is exactly the \textit{entropy} of the distribution as defined in \Cref{sec:probabilities}.

We substitute \Cref{eqn:kldivderivation5} back into \Cref{eqn:multikldiv1} to obtain that 
\begin{align}
    -{\nabla}_{\bm{\phi}} \sum_{d=1}^D D_{\text{KL}}(q_{\bm{\phi}}(\mathbf{z}^{(d)}|\mathbf{x})||p(\mathbf{z})) &= {\nabla}_{\bm{\phi}} \sum_{d=1}^D \left( - \log K + \text{Entropy}(f_{\bm{\phi}}(\mathbf{x})^{(d)}) \right) \\ 
    &= {\nabla}_{\bm{\phi}} \left( - D\log K + \sum_{d=1}^D\text{Entropy}(f_{\bm{\phi}}(\mathbf{x})^{(d)}) \right) \label{eqn:discreteKL2}\\
    &= {\nabla}_{\bm{\phi}} \sum_{d=1}^D\text{Entropy}(f_{\bm{\phi}}(\mathbf{x})^{(d)}) \label{eqn:discreteKL3} \\
    &= {\nabla}_{\bm{\phi}} \; \ov{\text{Entropy}}(f_{\bm{\phi}}(\mathbf{x})) \label{eqn:discreteKL4} \marginnote{\small \textcolor{\margincol}{Derived gradients w.r.t. $\bm{\phi}$ from  the $D_{\text{KL}}$ part of \cref{eqn:abstractphigrads}.}}[-0.3cm]
\end{align}
Note here that, although the $D\log K$ term is constant with respect to $\bm{\phi}$ and doesn't contribute to the \textit{gradient}, it will still contribute to the \textit{ELBO} and will be an important factor in comparing the (test set) performance across different VAE architectures that have differing latent dimension sizes. Line (\ref{eqn:discreteKL4}) follows the notation described in \Cref{eqn:aggEntropy}.

We are only left with calculating the gradients with respect to the remaining term on the right-hand side of \Cref{eqn:abstractphigrads}. Unlike the ${\nabla}_{\bm{\theta}}$ gradients, we \textit{cannot} apply Monte Carlo sampling here as the expectation is a function of $\bm{\phi}$ and thus
\begin{equation}
    {\nabla}_{\bm{\phi}} \mathbb{E}_{q_{\bm{\phi}}(\mathbf{z}|\mathbf{x})}\left[ \log p_{\bm{\theta}}(\mathbf{x}| \mathbf{z}) \right] \neq \mathbb{E}_{q_{\bm{\phi}}(\mathbf{z}|\mathbf{x})}\left[{\nabla}_{\bm{\phi}} \log p_{\bm{\theta}}(\mathbf{x}| \mathbf{z}) \right].
\end{equation}
In the more standard case of a VAE with a Gaussian distributed latent space, the so-called \textit{reparameterization trick} is typically used at this point as a solution. Unfortunately, this is not as straightforward in the discrete case. This is the key challenge of discrete VAEs, and a large body of literature exists attempting to solve this challenging optimization problem \citep[e.g.][]{bengio2013estimating,jang2017categorical,grathwohl2018backpropagation,liu2023bridging}. In our case, we will keep things as simple as possible and use the \textit{log-derivative trick}, allowing us to obtain samples that are unbiased estimates of the true gradient. A complete derivation is provided in \Cref{sec:logderivative}, but the key idea is to notice that
\begin{equation}
    {\nabla}_{\bm{\phi}} \mathbb{E}_{q_{\bm{\phi}}(\mathbf{z}|\mathbf{x})}\left[ \log p_{\bm{\theta}}(\mathbf{x}| \mathbf{z}) \right] = \mathbb{E}_{q_{\bm{\phi}}(\mathbf{z}|\mathbf{x})}\left[   \log p_{\bm{\theta}}(\mathbf{x}| \mathbf{z}) {\nabla}_{\bm{\phi}}\log q_{\bm{\phi}}(\mathbf{z}|\mathbf{x})\right].
\end{equation}
Now this expression \textit{can} be Monte Carlo sampled as the gradient is inside the expectation allowing us to obtain an unbiased (but potentially high variance) estimate of the gradient given by 
\begin{equation} \label{eqn:gradsphigeneralmc}
    {\nabla}_{\bm{\phi}} \mathbb{E}_{q_{\bm{\phi}}(\mathbf{z}|\mathbf{x})}\left[ \log p_{\bm{\theta}}(\mathbf{x}| \mathbf{z}) \right] \overset{\mathrm{mc}}{\approx} \log p_{\bm{\theta}}(\mathbf{x}| \mathbf{z}) {\nabla}_{\bm{\phi}} \log q_{\bm{\phi}}(\mathbf{z}|\mathbf{x}).
\end{equation}
For $\log p_{\bm{\theta}}(\mathbf{x}| \mathbf{z})$, we can repeat the steps used in \Cref{eqn:discretethetagrads3} to notice that
\begin{align} 
    \log p_{\bm{\theta}}(\mathbf{x}| \mathbf{z}) &= -\sum_{p=1}^P\text{BCE}(g_{\bm{\theta}}(\mathbf{z})^{(p)}, \mathbf{x}^{(p)}) \\
    &= - \; \ov{\text{BCE}}(g_{\bm{\theta}}(\mathbf{z}), \mathbf{x}).\label{eqn:lossphileft}
\end{align}
The $\log q_{\bm{\phi}}(\mathbf{z}|\mathbf{x})$ term is similar but each $\mathbf{z}^{(d)}$ has $K$ categories and therefore follows the more general categorical distribution. We begin by re-expressing in terms of the individual, categorically distributed latent neurons
\begin{align}
    \log q_{\bm{\phi}}(\mathbf{z}|\mathbf{x}) &= \log \prod_{d=1}^Dq_{\bm{\phi}}(\mathbf{z}^{(d)}|\mathbf{x}) \\
    &= \sum_{d=1}^D \log q_{\bm{\phi}}(\mathbf{z}^{(d)}|\mathbf{x})
\end{align}

As described in \Cref{eqn:categoricaldist}, the \textit{probability mass function} of the categorical distribution takes a similar form to the Bernoulli case,
\begin{equation}
    f_X(x) = \sum_{i=1}^k[x=k] \cdot p_k.
\end{equation}
We can substitute this back into the previous expression to inform us how it can be calculated.
\begin{equation}
\begin{split}
    \sum_{d=1}^D \log q_{\bm{\phi}}(\mathbf{z}^{(d)}|\mathbf{x}) = \sum_{d=1}^D \sum_{k=1}^K [\mathbf{z}^{(d)}_k = k] \log q_{\bm{\phi}}(\mathbf{z}^{(d)}_k = k|\mathbf{x}).
\end{split}
\end{equation}
Although this may appear complex, this is really just the sum of the model's probabilities of the categories that were actually selected during the sampling process $\mathbf{z} \sim \text{Cat}$. To simplify, we can write this in terms of the encoder network $f_{\bm{\phi}}(\mathbf{x})$ and simply index the appropriate outputs. Let us denote the index of the sampled category for latent variable $d$ as $k(d)$, then we can write the previous equation as
\begin{equation} \label{eqn:lossphiright}
    \sum_{d=1}^D \log q_{\bm{\phi}}(\mathbf{z}^{(d)}|\mathbf{x}) = \sum_{d=1}^D \log f_{\bm{\phi}}(\mathbf{x})^{(d)}_{k(d)}.
\end{equation}
Substituting \Cref{eqn:lossphileft} and \Cref{eqn:lossphiright} back into \Cref{eqn:gradsphigeneralmc} we obtain our final expression for the second term in the gradients of $\bm{\phi}$ giving
\begin{equation}
    {\nabla}_{\bm{\phi}} \mathbb{E}_{q_{\bm{\phi}}(\mathbf{z}|\mathbf{x})}\left[ \log p_{\bm{\theta}}(\mathbf{x}| \mathbf{z}) \right]   \overset{\mathrm{mc}}{\approx} - \;\ov{\text{BCE}}(g_{\bm{\theta}}(\mathbf{z}), \mathbf{x}) \left( {\nabla}_{\bm{\phi}} \sum_{d=1}^D \log f_{\bm{\phi}}(\mathbf{x})^{(d)}_{k(d)} \right) . \label{eqn:discretphigradsrhs}\marginnote{\small \textcolor{\margincol}{Derived gradients w.r.t. $\bm{\phi}$ from  the $\mathbb{E}[\,\cdot\,]$ part of \cref{eqn:abstractphigrads}.}}[-0.3cm]
\end{equation}

Finally, we are likely going to want to track the performance of the discrete VAE via the $\mathcal{L}^{\text{ELBO}}_{\bm{\theta}, \bm{\phi}}(\mathbf{x})$ term. This will allow us to, for example, determine when the model is finished training or compare performance across different architectures. Thankfully, the hard work is already done for this expression and we can simply revisit our derivation of \Cref{eqn:elbodecomp4} by revisiting our workings for obtaining \Cref{eqn:abstractphigrads} prior to calculating the gradients with respect to $\bm{\phi}$. Specifically, our estimate of the ELBO will take the following form:

\begin{align}
 \mathcal{L}^{\text{ELBO}}_{\bm{\theta}, \bm{\phi}}(\mathbf{x}) &= - D_{\text{KL}}(q_{\bm{\phi}}(\mathbf{z}|\mathbf{x})||p(\mathbf{z})) + \mathbb{E}_{q_{\bm{\phi}}(\mathbf{z}|\mathbf{x})}\left[ \log p_{\bm{\theta}}(\mathbf{x}| \mathbf{z}) \right] \\
 &\overset{\mathrm{mc}}{\approx} \underbrace{\ov{\text{Entropy}}(f_{\bm{\phi}}\left(\mathbf{x})\right) - D\log K}_{\Cref{eqn:discreteKL2}} - \underbrace{\ov{\text{BCE}}(g_{\bm{\theta}}(\mathbf{z}), \mathbf{x}).}_{ \Cref{eqn:discretethetagrads4}} \marginnote{\small \textcolor{\margincol}{Calculation of the ELBO from \cref{eqn:ELBO}.}}[-0.2cm]
\end{align}
All that is left now is to put everything we have derived together and to train our discrete VAE!  

\section{Summarizing the Discrete VAE} \label{sec:summary}
The previous section derived everything we need to train a discrete VAE. Now we can simply summarize the concrete steps required for a practical implementation. We can implement the encoder $f_{\bm{\phi}}(\mathbf{x})$ and decoder $g_{\bm{\theta}}(\mathbf{z})$ using any neural network architecture we like. Then we will apply standard stochastic gradient ascent across some training dataset where the gradients for each training example are given by 
\begin{align}
    \hat{\nabla}_{\bm{\phi}} &\overset{\mathrm{mc}}{\approx}\underbrace{\vphantom{\sum_{d=1}^D}{\nabla}_{\bm{\phi}} \; \ov{\text{Entropy}}(f_{\bm{\phi}}(\mathbf{x}))}_{\Cref{eqn:discreteKL4}} \; - \; \underbrace{\ov{\text{BCE}}(g_{\bm{\theta}}(\mathbf{z}), \mathbf{x}) \left( {\nabla}_{\bm{\phi}} \sum_{d=1}^D \log f_{\bm{\phi}}(\mathbf{x})^{(d)}_{k(d)} \right)}_{\Cref{eqn:discretphigradsrhs}} \label{eqn:summarygradphi} \\
    \hat{\nabla}_{\bm{\theta}} &\overset{\mathrm{mc}}{\approx} \underbrace{-{\nabla}_{\bm{\theta}} \; \ov{\text{BCE}}(g_{\bm{\theta}}(\mathbf{z}), \mathbf{x})}_{\Cref{eqn:discretethetagrads3}}. \label{eqn:summarygradtheta}
\end{align}
As we have shown, by using these gradients we are maximizing the ELBO which is the core idea of the VAE as it maximizes the log-likelihood of the data under our model $\log p_{\bm{\theta}}(\mathbf{x})$ \textit{and} minimizes the divergence between our model of the posterior $q_{\bm{\phi}}(\mathbf{z}| \mathbf{x})$, and the problematic target posterior $p_{\bm{\theta}}(\mathbf{z} | \mathbf{x})$. We can also calculate our estimate of the value of the ELBO for a given sample as
\begin{align}
 \mathcal{L}^{\text{ELBO}}_{\bm{\theta}, \bm{\phi}}(\mathbf{x})
 \overset{\mathrm{mc}}{\approx} \ov{\text{Entropy}}(f_{\bm{\phi}}(\mathbf{x})) - D\log K - \ov{\text{BCE}}(g_{\bm{\theta}}(\mathbf{z}), \mathbf{x}).
\end{align}
For completeness, we include the training recipe for a discrete VAE in \Cref{alg:discreteVAE}.
\begin{algorithm}
\caption{Discrete VAE with Stochastic Gradient Ascent}
\label{alg:discreteVAE}
\begin{algorithmic}
\item \textbf{Input: } $\textrm{Encoder } f_{\bm{\phi}}, \textrm{decoder } g_{\bm{\theta}}, \textrm{training data } \mathcal{D}^\text{train}, \textrm{validation data } \mathcal{D}^\text{val},  \textrm{learning rate } \eta$;
\item \textbf{Initialize: } $\bm{\phi}$ \& $\bm{\theta}$;
\While{$\mathcal{L}^{\text{ELBO}}_{\bm{\theta}, \bm{\phi}}(\mathbf{\mathcal{D}^\text{val}})$ not converged}
        \State Sample batch $\bm{x} \sim \mathcal{D}^\text{train}$ without replacement; 
        \State Sample latent $\mathbf{z} \sim f_{\bm{\phi}}(\mathbf{x})$; \Comment{As described in \Cref{sec:thediscretevae}}
        \State Decode latents $\hat{\mathbf{x}} = g_{\bm{\theta}}(\mathbf{z})$
        \State Update encoder $\bm{\phi} \leftarrow \bm{\phi} + \eta \cdot \hat{\nabla}_{\bm{\phi}}$; \Comment{Using \Cref{eqn:summarygradphi}}
        \State Update decoder $\bm{\theta} \leftarrow \bm{\theta} + \eta \cdot \hat{\nabla}_{\bm{\theta}}$; \Comment{Using \Cref{eqn:summarygradtheta}}
      \EndWhile
\end{algorithmic}
\end{algorithm}

This algorithm can be easily implemented using any automatic differentiation framework which greatly simplifies the calculation of gradients required in \Cref{eqn:summarygradphi,eqn:summarygradtheta}. We include a simple example of such an implementation using the \texttt{PyTorch} library at \url{www.github.com/alanjeffares/discreteVAE}.

\newpage

\newpage
\bibliographystyle{alpha}
\bibliography{references}


\end{document}